%% file: main.tex
\PassOptionsToPackage{prologue, dvipsnames, svgnames}{xcolor}
\documentclass[10pt, a4paper]{article}

\usepackage[]{lrec-coling2024} 
\usepackage{array}
\usepackage{caption}
\usepackage{multirow}
\usepackage{graphicx}
\usepackage{array}
\usepackage{enumitem}
\newcommand{\beq}[1]{\vspace{-0.03in}\begin{equation}#1\end{equation}\vspace{-0.03in}}

\newcommand{\hide}[1]{}
\usepackage{amsmath, bm}
\usepackage{balance}
\newcommand{\vpara}[1]{\vspace{0.05in}\noindent \textbf{#1 }}
\newcommand{\tabincell}[2]{\begin{tabular}{@{}#1@{}}#2\end{tabular}}

\usepackage[linesnumbered,ruled,vlined]{algorithm2e}
\usepackage{algorithmic}
\usepackage{longtable}
\usepackage{booktabs}
\usepackage{threeparttable}

\title{Diversifying Question Generation over Knowledge Base via External Natural Questions}

\name{Shasha Guo\textsuperscript{1, 2}, Jing Zhang\textsuperscript{1, 2 *}\thanks{\text{*}\scriptsize Corresponding author.\normalsize}, Xirui Ke\textsuperscript{1, 2}, Cuiping Li\textsuperscript{1, 2}, Hong Chen\textsuperscript{1, 2}
}

\address{\textsuperscript{1}School of Information, Renmin University of China, Beijing, China\\
  \textsuperscript{2}Key Laboratory of Data Engineering and Knowledge Engineering of Ministry of Education\\
  \{guoshashaxing, zhang-jing, kexirui, licuiping, chong\}@ruc.edu.cn\\}

\abstract{
Previous methods on knowledge base question generation (KBQG) primarily focus on refining the quality of a single generated question.
However, considering the remarkable paraphrasing ability of humans, we believe that diverse texts can express identical semantics through varied expressions.
The above insights make diversifying question generation an intriguing task, where the first challenge is evaluation metrics for diversity.
Current metrics inadequately assess the aforementioned diversity. They calculate the ratio of unique n-grams in the generated question, which tends to measure duplication rather than true diversity.
Accordingly, we devise a new diversity evaluation metric, which measures the diversity among top-k generated questions for each instance while ensuring their relevance to the ground truth.
Clearly, the second challenge is how to enhance diversifying question generation.
To address this challenge, we introduce a dual model framework interwoven by two selection strategies to generate diverse questions leveraging external natural questions. 
The main idea of our dual framework is to extract more diverse expressions and integrate them into the generation model to enhance diversifying question generation.
Extensive experiments on widely used benchmarks for KBQG show that our approach can outperform pre-trained language model baselines and text-davinci-003 in diversity while achieving comparable performance with ChatGPT.
 \\ \newline  }

\begin{document}

\maketitleabstract

\input{intro.tex}

\input{evaluation.tex}

\input{method.tex}

\input{experiment.tex}

\input{related.tex}

\input{conclusion.tex}

\input{limitations}
\input{ack}
\section{Bibliographical References}
\label{sec:reference}
\bibliographystyle{lrec-coling2024-natbib}
\bibliography{sample-base}

\appendix
\input{appendix}

\end{document}

%% file: intro.tex
\section{Introduction}
\label{sec:intro1}
Knowledge Base Question Generation (KBQG) is an essential task, which focuses on generating natural language questions based on a set of formatted facts extracted from a knowledge base (KB).
In recent years, KBQG has attracted substantial research interest due to its wide range of applications.
For example, in education, KBQG can generate numerous questions from course materials, aiding in assessing students' grasp of the content and enhancing self-learning
~\cite {MathWordProblem, MWPG}. Furthermore, in industry, KBQG can encourage machines to actively ask human-like questions in human-machine conversations~\cite{DialogueSystem, zeng2020exploiting}.
Additionally, KBQG can augment training data to boost the quality of question answering (QA) tasks~\cite{chen2020toward, DSM}.

\begin{figure}[!t]
\centering 
\includegraphics[width=0.48\textwidth]{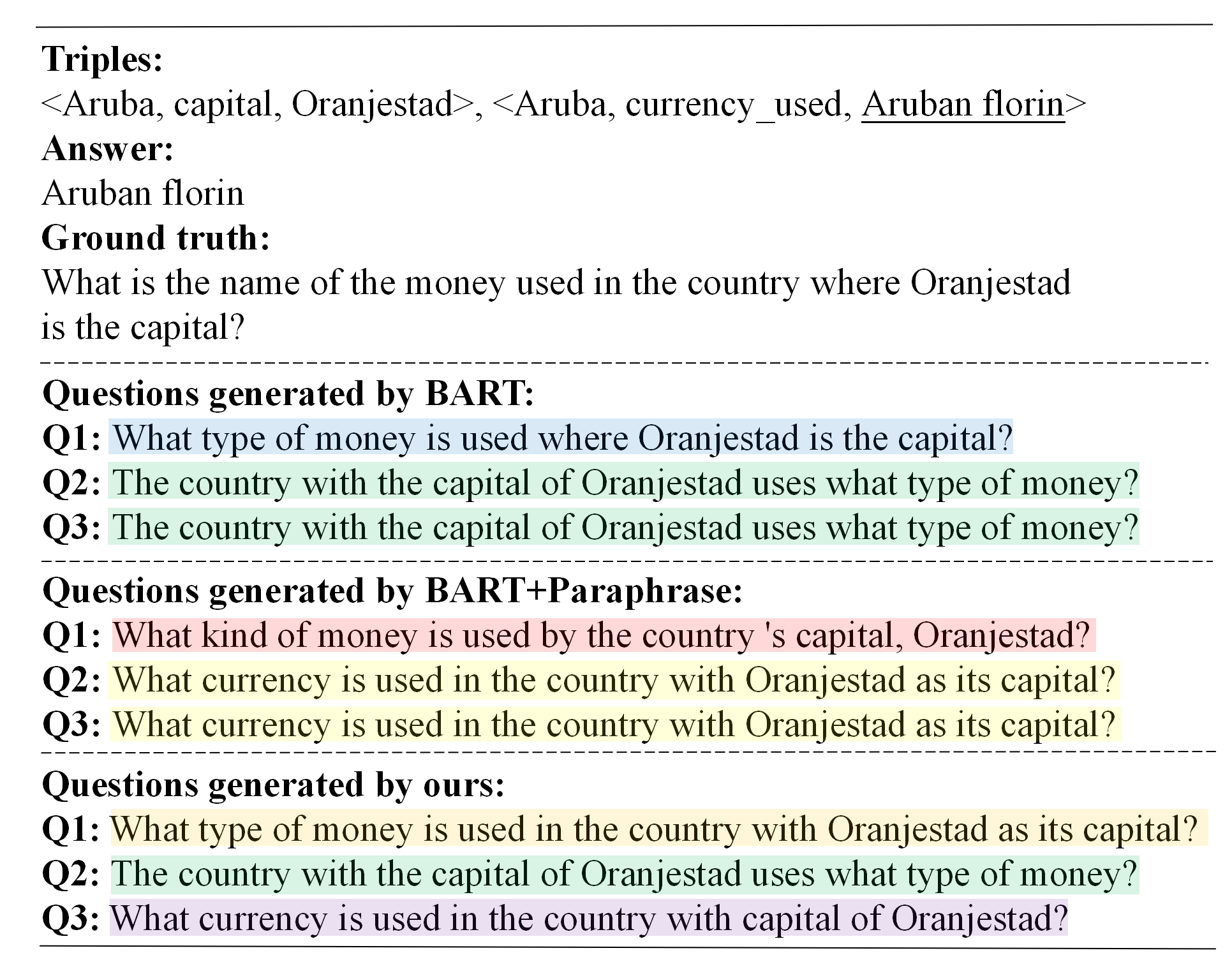}
\caption{Example questions generated by BART, BART+Paraphrase, and our approach on the WebQuestions (WQ) dataset. Given a set of triples with the answer(underlined), each method returns top-3 questions, where the various surface forms are marked in different colors.}
\label{fig:example} 
\end{figure}

Recently, pre-trained language models (PLMs)-based methods~\cite{KGPT, JointGT, AutoQGS} have achieved advanced performance on KBQG. 
Despite the success of these models, they conform to the one-to-one encoder-decoder paradigm and concentrate on improving the quality of the generated question, resulting in insufficient diversity.
In fact, human communication exhibits a remarkable ability to paraphrase, which means humans can express the same semantics in different surface forms, such as words, phrases, and grammatical patterns.
Figure~\ref{fig:example} gives an illustration of several diverse questions (those generated by ours), expressing the same semantics from the identical input triples of KBs. 
Intuitively, we think that the diversity of texts should be that texts expressing the same semantics have different forms of expression.

However, current evaluation metrics about relevance and diversity deviate from the above observation. 
To measure the relevance between the generated question and the ground truth, BLEU~\cite{papineni2002bleu} and ROUGE~\cite{rouge} have been proposed, which simply calculate the ratio of common n-grams in two texts without considering semantics. 
Unlike the above metrics for computing n-grams similarity, BERTScore~\cite{BERTScore} further measures token-level semantic similarity between two texts. Nonetheless, we believe that sentence-level semantic similarity between the two texts is more important.
As for the diversity of the generated question, metrics like Distinct-n~\cite{Distinct} measure the percentage of unique n-grams within the question itself.
It can be viewed as a measure of the duplication of n-grams in the generated question, which does not conform with the diversity defined above.

Towards these evaluation metrics, existing PLMs-based models fail to mimic humans to produce diverse questions. 
For one thing, the metrics for assessing diversity are inappropriate.
For another, these models strive to make the generated question similar to the ground truth question, which limits and narrows down the search space when decoding the output.
Therefore, a natural solution is to expand the search space by increasing the ground truth question.
To illustrate such an idea, we first perform a pilot study (Cf. the detailed settings in Section~\ref{sec: pilot study}) by conducting one preliminary experiment that augments the ground truth questions with automatically paraphrased questions.
We make the observation that \textbf{injecting paraphrased questions results in a more diverse set of generated questions.}
However, due to the limited capability of the paraphrase model, the paraphrased questions exhibit only slight vocabulary variations compared to the ground truth question, which leaves the potential for further exploration.

Inspired by the above insights, in this paper, we propose a novel diversity evaluation metric called $Diverse@k$, which measures the diversity among the top-k generated questions for each instance while ensuring their relevance to the ground truth. 
Additionally, we investigate the use of a wide range of external natural questions~\cite{NaturalQuestions} to enhance the diversity of question generation.
We believe that large-scale external corpus, instead of being restricted to a small amount of training data, can provide a wider and more diverse range of linguistic and semantic expressions.

To extract rich expressions and squeeze them into the generation model, we design two dual models, namely the forward model and the backward model, to transform the formatted facts like triples into the natural question and vice versa respectively. 
We further design two simple yet effective selection strategies of pseudo pairs to interweave the two models.
The first selection strategy is imposed on the output of the backward model. 
Given an external question, it is input into the backward model to generate the corresponding triples. By evaluating the likelihood of the triples generating the external question in the forward model, reliable and high-quality pseudo-pairs are identified and filtered, thereby enhancing the diversity of the training data.
The second strategy is applied to the output of the forward model. 
Given the triples from the training data and the generated questions, semantic relevance and diversity scores are used together to sift out similar but different questions for each instance, improving the capacity of the backward model for dealing with the external questions.
As a result, semantically similar yet more linguistically diverse natural questions are generated by iteratively flowing pseudo-data between the two models, providing substantial advantages over traditional paraphrased methods.

\vpara{Contributions.} (1) To the best of our knowledge, we are the first to propose the diversity among top-k generated questions for each instance, ensuring their relevance to the ground truth, and design a novel metric to measure the diversity. (2) We present a dual model framework interwoven by two selection strategies to assemble a variety of diverse questions from external natural questions, enabling diverse expressions to be injected into the generation model. 
(3) Extensive experiments show our model consistently exhibits superior diversity. It surpasses pre-trained language model baselines and text-davinci-003~\cite{InstructGPT}, while achieving comparable performance with ChatGPT\footnote{https://openai.com/blog/chatgpt}.

%% file: evaluation.tex
\section{Rethinking Diversity Evaluation}
\label{sec:evaluation}
A metric is essential to evaluate a generation model's capacity to produce diverse questions from identical input, as illustrated in Figure~\ref{fig:example}.

\vpara{Distinct-n.}
Previous works~\cite{www21, Dialog1} mostly use $Distinct\text{-}n$~\cite{Distinct}, \emph{i.e.}, $Distinct\text{-}n = \frac{| unique \ n\text{-}grams |}{| total \ n\text{-}grams |}$, to calculate the diversity score of the generated text. 
Some works assess $Distinct\text{-}n$ in instance-level~\cite{AskGoodQuestion, www21}, while others treat all instances as a whole to compute the unique n-grams in the total n-grams of all instances~\cite{Dialog1, Dialog2}. Clearly, neither of them is appropriate, as $Distinct\text{-}n$ actually focuses on the proportion of unique n-grams, which appears more akin to evaluating duplication than diversity.

\vpara{Diverse@k.} 
We propose $Diverse@k$ as a new metric to assess the diversity of the top-k generated questions for each instance.
The main idea is to sum the pairwise diversity of the top-k generated questions.
Specifically, given two generated questions $S_i$ and $S_j$ with the ground truth question $S$, $Diverse@k$ is defined as:

\beq{
\label{eq:diverse}
\begin{split}
        Diverse@k = \sum_{i=1}^{k-1} \sum_{j= i+1}^{k} Diverse(S_i, S_j),\\
     Diverse(S_i, S_j) = \frac{|\mathcal{T}_i - \mathcal{T}_j|+|\mathcal{T}_j - \mathcal{T}_i|}{|\mathcal{T}_i \cup \mathcal{T}_j| }, \\
      R(S_i, S) \geq \alpha \text{ and} \ R(S_j, S) \geq \alpha
\end{split}  
}

\noindent where $\mathcal{T}_i$ and $\mathcal{T}_j$ are the sets of tokens in $S_i$ and $S_j$, respectively, so $Diverse(S_i,S_j)$ measures their differences. 
Then we sum $Diverse(S_i, 
 S_j)$ of all pairwise top-k generated questions to represent the diversity of the instance. Clearly, mere summation fails to accurately reflect the quality of the generated questions.
Consequently, we impose constraints on semantic similarity to guarantee the relevance of each generated question.
Specifically, we use simCSE~\cite{SimCSE}, a popular semantic relevance metric, to measure the relevance score between $S_i$ and the ground truth question $S$ and denote it as $R(S_i, S)$. $\alpha$ is the threshold of the relevance score. We set it as \textbf{70\%} (Cf. the detailed explanation in Section~\ref{exp:overall}(3) ) to filter out irrelevant questions.
 Moreover, we evaluate the correlation between $Diverse@k$ and human evaluation using the Pearson correlation coefficient. Table~\ref{tb:human_evaluation} reports the results on $Diverse@3$ (\emph{i.e.}, \textbf{0.921})  and $Diverse@5$ (\emph{i.e.}, \textbf{0.935}) respectively.

%% file: method.tex
\begin{figure*}[!t]
\centering 
\includegraphics[width=0.95\linewidth]{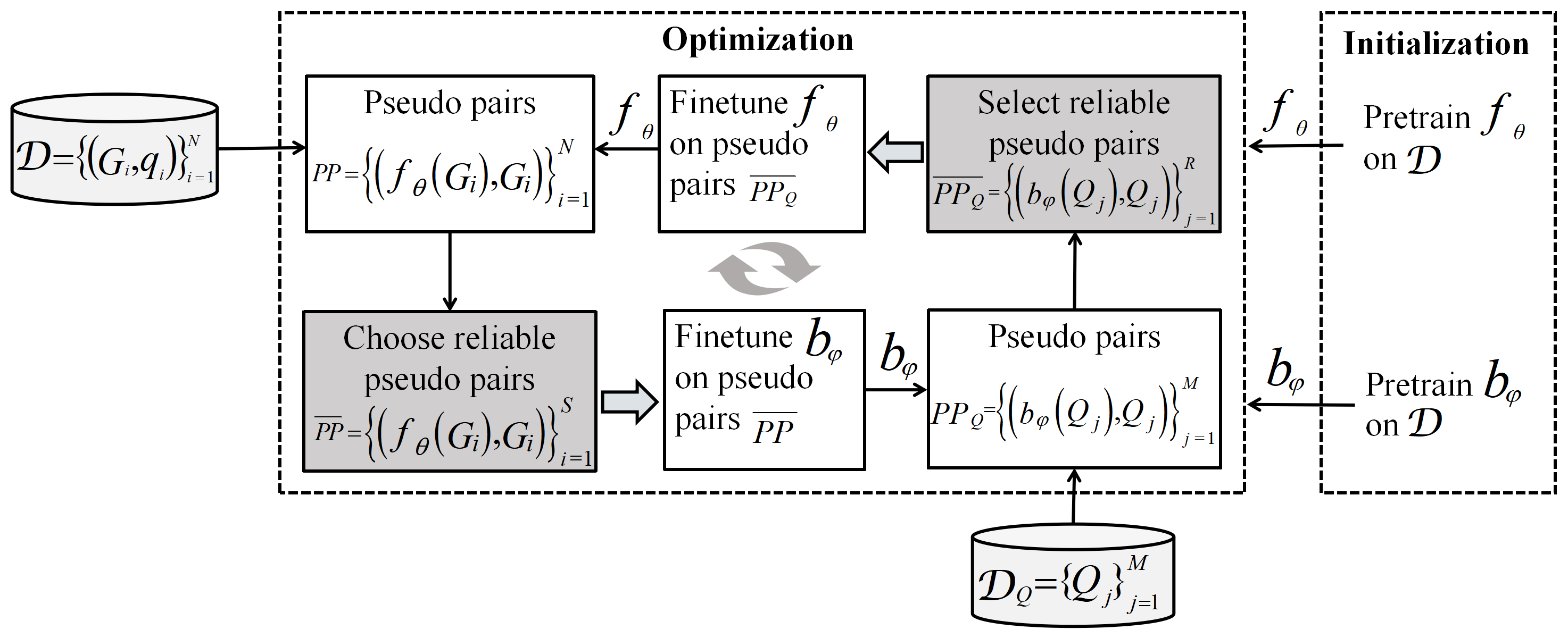}
\caption{Overview of our proposed method.
The forward model $f_\theta$ and the backward model $b_\varphi$ are pre-trained on training data $\mathcal{D}$ and then iteratively fine-tuned on reliable pseudo pairs $\{(b_{\varphi}(Q_j), Q_j) \}_{j=1}^R$ and $\{(f_{\theta}(G_i), G_i )\}_{i=1}^S$ respectively, which are filtered by our proposed selection strategies.   
}
\label{fig:overview}
\end{figure*}

\section{Approach}
\label{sec:method}

\subsection{Problem Definition}
\label{sec:problem define}
KBQG aims to generate questions given a set of triple facts represented as a subgraph.
Formally, given a dataset $\mathcal{D} = \{(G_i, q_i)\}_{i=1}^N$, 
where $G_i$ represents a subgraph consisting of a set of connected triples and $q_i$ signifies the corresponding question, the objective of KBQG is to learn a function $f$ with parameter $\theta$ to map from $G_i$ to $q_i$, \emph{i.e.}, $f_{\theta}: G_{i} \xrightarrow[]{} {q}_i$.

\subsection{Pilot Study}
\label{sec: pilot study}
We conduct a preliminary experiment by paraphrasing target questions to show their positive effect on diversifying question generation.

\vpara{Modeling.}
We employ t5-large-paraphraser-diverse-high-quality\footnote{https://huggingface.co/ramsrigouthamg/t5-large-paraphraser-diverse-high-quality}, an advanced paraphrase model to paraphrase ground truth questions in the training data automatically.
We create three paraphrased questions $(q_{i}^{1}, q_{i}^{2}, q_{i}^{3})$ for each ground truth question $q_i$.
We use BART-base\footnote{https://huggingface.co/facebook/bart-base}~\cite{BART} as the backbone of our generation model, as prior research~\cite{chen2020toward} has demonstrated that BART results in state-of-the-art question generation performance. 
We fine-tune BART on the paraphrased dataset and denote the method as BART+Paraphrase (abbreviated as B+P).

\vpara{Observation.} Table~\ref{tab:pilot} illustrates the evaluation results measured by our proposed metric $Diverse@k$ (\emph{i.e.}, $Diverse@3$) .
The results show that questions generated by injecting paraphrased ground truth are more diverse than those generated solely from the original ground truth, indicating that paraphrasing has a positive effect on enhancing diversifying question generation.
Since these paraphrased questions contain much richer semantic patterns and expressions than the ground truth, the generation model can learn from them to obtain more diverse question expressions.

\begin{table}[!t]
\centering
\newcolumntype{?}{!{\vrule width 1pt}}
\renewcommand\arraystretch{1.0}
\captionsetup{font=footnotesize}
\begin{tabular}{c?c}
\toprule
\textbf{Model}
& \textbf{Diverse@3}\\
\midrule
 BART & 18.98 \\
BART+Paraphrase & \textbf{21.52} \\
\midrule
Gain & \textbf{2.54} \\
\bottomrule
\end{tabular}
\caption{Performance comparison between BART and BART+Paraphrase for KBQG on the PQ dataset (\%).}
\label{tab:pilot}
\end{table}

\vpara{Discussion.} Above we propose a simple but
effective approach to construct one-to-many instances to expand the searching space of the generation model. 
Despite the advantages of these paraphrased questions, they only exhibit minor differences from the target questions and are limited in scale due to the inadequate capacity of the paraphrase model.
A straightforward method is to make efforts to design promising paraphrase models, but we explore another way to acquire diversity by leveraging external natural questions.
Compared with paraphrased questions, external natural questions can cover a much broader range of semantic patterns and language expressions.
Moreover, natural questions are more human-like, while paraphrased questions are relatively rigid and mechanical.
In view of this, we attempt to employ external natural questions to diversify question generation.

\subsection{Model Overview}
\label{sec:overview}
In this work, we leverage external natural questions denoted as $\mathcal{D_Q} = \{Q_j\}_{j=1}^M$ to diversify question generation, where $Q_j$ has no corresponding subgraph. 
Figure~\ref{fig:overview} illustrates the overview of our proposed approach.
At a high level, our approach consists of two steps including \textbf{initialization} and \textbf{optimization}.
Motivated by the concept of back translation~\cite{MTACL22, machine}, we propose a backward model to help the forward model capture diverse question expressions.
First, the forward model $f_\theta$ and the backward model $b_\varphi$ are pre-trained on the training data $\mathcal{D}$ to obtain a good initialization point.
Next, we employ $\mathcal{D_Q}$ to optimize $f_\theta$ and $b_\varphi$ simultaneously. 
Specifically, we use $b_\varphi$ to generate corresponding triple-based sequences of $\mathcal{D_Q}$ to train $f_\theta$. By doing this, $f_\theta$ can capture more diverse expressions from $\mathcal{D_Q}$.
Then, we use the above optimized $f_\theta$ to generate questions for the triple-based sequences in training data $\mathcal{D}$, which are then used to train $b_\varphi$.
By doing this, the pseudo questions of the training data $\mathcal{D}$ can provide $b_\varphi$ with more diverse expressions than the ground truth. 
Thus, the optimized $b_\varphi$ can deal with various external questions to further offer more diverse pseudo data for $f_\theta$.
With continuous iterative training, $f_\theta$ and $b_\varphi$ constantly gather substantial diverse questions for the training data, which are subsequently injected into $f_\theta$ along with each iteration to acquire diversity.

\subsection{Step 1: Initialization}
\label{sec:step1}
We use BART~\cite{BART} to instantiate the forward model $f_{\theta}$.
Concretely, we linearize each subgraph $G_i$ into a triple-based sequence,
where each triple is separated by the special token ``$<$/s$>$''. Then, we input the sequence into BART to generate a corresponding question.
We finally fine-tune $f_{\theta}$ on $\mathcal{D}$ by maximizing the probabilities of generating all gold questions, \emph{i.e.},

\beq{
\label{eq:forward}
    \mathcal{L}_{f}^{(0)} = \max_{\theta} \sum_{i=1}^{N} \log P_{\theta}\left(q_i| G_i\right)
}

Similarly, we also employ BART to instantiate the backward model $b_{\varphi}$. 
Specifically, each question $q_i$ is fed into $b_{\varphi}$ to generate a triple-based sequence.
Then, we fine-tune $b_{\varphi}$ on $\mathcal{D}$ by maximizing the probabilities of generating all gold triple-based sequences, \emph{i.e.},

\beq{
\label{eq:backward}
    \mathcal{L}_{b}^{(0)} = \max_{\varphi} \sum_{i=1}^{N} \log P_{\varphi}\left(G_i| q_i\right)
}

\subsection{Step 2: Optimization}
\label{sec:step2}
In this section, we explain how to iteratively fine-tune $f_\theta$ and $b_\varphi$ on  $\mathcal{D_Q}$ and $\mathcal{D}$. Through this iterative fine-tuning, $f_\theta$ and $b_\varphi$ accumulate a wide range of diverse patterns and expressions from $\mathcal{D_Q}$, which endows the forward model with diversity. 

\vpara{Forward Model.}
We explain how to fine-tune the forward model $f_{\theta}$ with external natural questions $\mathcal{D_Q} = \{Q_j\}_{j=1}^M$.
Because questions in $\mathcal{D_Q}$ do not have corresponding triple-based sequences, we use the backward model $b_{\varphi}$ to generate them and construct the pseudo pairs $\{(b_{\varphi}(Q_j), Q_j) \}_{j=1}^M$. Then, the forward model $f_{\theta}$ is fine-tuned on these pseudo pairs by maximizing the probabilities of generating the external questions, \emph{i.e.},

\beq{
\label{eq:forward_train}
    \mathcal{L}_{f} = \max_{\theta} \sum_{j=1}^{M} \log P_{\theta}\left(Q_j|b_{\varphi}(Q_j)\right)
}

\noindent where $b_{\varphi}(Q_j)$ is an abbreviation for $b_{\varphi}(G_i | Q_j)$. 
Since $b_{\varphi}$ is pre-trained on training data $\mathcal{D}$, $b_{\varphi}(Q_j)$ generally follows the patterns of triple-based sequences in 
 $\mathcal{D}$.
External questions $\mathcal{D_Q}$ provide more semantic patterns and expressions, enabling $f_{\theta}$ to generate more diverse questions.

\vpara{Backward Model.}
We elaborate how to fine-tune the backward model $b_{\varphi}$ based on training data $\mathcal{D} = \{(G_i, q_i)\}_{i=1}^N$.
Beyond the gold question $q_i$ of $G_i$, we employ the forward model $f_{\theta}$ to generate the new question, namely $f_{\theta}(G_i)$.
Because $f_{\theta}$ is fine-tuned on external natural questions $\mathcal{D_Q}$, $f_{\theta}(G_i)$ could have different expressions from ground truth $q_i$.
We organize $f_{\theta}(G_i)$ and $G_i$ into the pseudo pairs $\{(f_{\theta}(G_i), G_i)\}_{i=1}^N$.
Then, the backward model $b_{\varphi}$ is fine-tuned on these pseudo pairs by maximizing the probabilities of generating all gold triple-based sequences, \emph{i.e.},

\beq{
\label{eq:backward_train}
    \mathcal{L}_{b} = \max_{\varphi} \sum_{i=1}^{N} \log P_{\varphi}\left(G_i|f_{\theta}(G_i)\right)
}

\noindent where $f_{\theta}(G_i)$ is an abbreviation for $f_{\theta}(q_i | G_i)$. Various $f_{\theta}(G_i)$ empower $b_{\varphi}$ the ability to deal with a variety of external questions.

\subsection{Reliable Pseudo Pairs Selection}
\label{sec:select}
During optimization, the forward model and the backward model are fine-tuned with pseudo pairs. 
To better assemble reliable diverse expressions for the training data,
we propose two simple yet effective selection strategies. 

First, we aim to extract diverse expressions from external questions to enrich the training data.
To achieve this goal, this selection designs the first likelihood-based selection strategy for filtering pseudo-pairs $\{(b_{\varphi}(Q_j), Q_j )\}_{j=1}^M$ generated by the backward model $b_{\varphi}$. Specifically, the reliability of the pseudo-pairs is evaluated based on the likelihood of the forward model  $f_\theta$ generating $\hat{Q}_j$ when $b_{\varphi}(Q_j)$ is used as input. A higher likelihood indicates that $b_{\varphi}(Q_j)$  is more reliable, as it receives a high-confidence score from the forward model. This strategy effectively selects high-quality pseudo-pairs, thereby enhancing the diversity and representational capacity of the training data used to train the forward model.

Subsequently, we need to further augment the extraction patterns for the backward model to enlarge the scale of selected external questions in subsequent steps.
Therefore, to expand the search space of $b_{\varphi}$, we design the second selection strategy based on both SimCSE~\cite{SimCSE} and our designed $Diverse(S_i,S_j)$. 
Specifically, for the pseudo pairs $\{(f_{\theta}(G_i), G_i) \}_{i=1}^N$ generated by the forward model, we first use simCSE to compute the relevance between $f_{\theta}(G_i)$ and $q_i$. 
This allows us to retain $f_{\theta}(G_i)$ with high semantic relevance, effectively filtering out noisy and irrelevant generated questions.
Then, we calculate the diverse score $Diverse(f_{\theta}(G_i), q_i)$ to select the pseudo pair $(f_{\theta}(G_i), G_i)$ with a higher score for fine-tuning the backward model.
The forward model provides a more varied expression for the formatted facts in the training data, indicating that the backward model can be fine-tuned to handle more external inquiries.

\begin{algorithm}[t]	\renewcommand{\algorithmicrequire}{\textbf{Input:}}
	\renewcommand{\algorithmicensure}{\textbf{Output:}}
	\renewcommand{\algorithmicreturn}{\textbf{Return}}
	\caption{\textbf{Our Proposed Approach}}
	\label{alg:our proposed}
	\begin{algorithmic}[1]
	    \REQUIRE $\mathcal{D} = \{(G_i, q_i)\}_{i=1}^N$, $\mathcal{D_Q} = \{Q_j\}_{j=1}^M$.
	    \ENSURE $\theta$ of the forward model $f_{\theta}$, $\varphi$ of the backward model $b_{\varphi}$.
        \STATE Instantiate $f_{\theta}$ and $b_{\varphi}$ via BART;
        \STATE Pretrain  $f_{\theta}$ on $\mathcal{D}$ via Eq.~\eqref{eq:forward};
        \STATE Pretrain  $b_{\varphi}$ on  $\mathcal{D}$ via Eq.~\eqref{eq:backward};
	\FOR{each iteration}
             \FOR{each epoch}
             \STATE Generate $M$ pseudo pairs $\{(b_{\varphi}(Q_j), Q_j) \}_{j=1}^M$;
	  \STATE  Generate $M$ pseudo questions $\{\hat{Q}_j\}_{j=1}^M$ = $\{f_{\theta}(b_{\varphi}(Q_j))\}_{j=1}^M$ on $\{b_{\varphi}(Q_j) \}_{j=1}^M$;    
   \STATE Select $R$ reliable pseudo pairs $\{(b_{\varphi}(Q_j), Q_j) \}_{j=1}^R$ if the likelihood of generating $\hat{Q}_j$ exceeds 60\%;
        \STATE Update $\theta$ based on $\{(b_{\varphi}(Q_j), Q_j) \}_{j=1}^R$ via Eq.~\eqref{eq:forward_train};
	    \ENDFOR
           \IF{iteration = last iteration}
          \STATE Finetune  $f_{\theta}$ on $\mathcal{D}$ via Eq.~\eqref{eq:forward};
        \ENDIF
              \FOR{each epoch}
	    \STATE Generate $N$ pseudo pairs $\{(f_{\theta}(G_i), G_i) \}_{i=1}^N$;
     \STATE Choose $S$ reliable pseudo pairs $\{(f_{\theta}(G_i), G_i)\}_{i=1}^S$ on semantic relevance and diversity between $f_{\theta}(G_i)$ and $q_i$;
	       \STATE Optimize $\varphi$ based on $\{(f_{\theta}(G_i), G_i) \}_{i=1}^S$ via Eq.~\eqref{eq:backward_train};
	    \ENDFOR
     
    \ENDFOR
     \STATE Return $\theta$ and $\varphi$.
        	\end{algorithmic}  
\end{algorithm}

The two selection strategies encourage two dual models to benefit by associating together and complementing each other.
We summarize the whole procedure by Algorithm~\ref{alg:our proposed}. In it, line 1 initializes the forward model $f_{\theta}$ and the backward model $b_{\varphi}$ via BART. Lines 2-3 pretrain $f_{\theta}$ and $b_{\varphi}$ based on $\mathcal{D}$ to get a good initialization point. Lines 4-19 iteratively train  $f_{\theta}$ and $b_{\varphi}$ util convergence. 
Line 6 generates the pseudo triple-based sequence $b_\varphi(Q_j)$ about $Q_j$ using $b_{\varphi}$, and constructs $M$ pseudo pairs $\{(b_{\varphi}(Q_j), Q_j) \}_{j=1}^M$. 
Line 7 adopts $f_\theta$ to generate $M$ pseudo questions $\{\hat{Q}_j\}_{j=1}^M$ on $\{b_\varphi(Q_j)\}_{j=1}^M$.
Line 8 chooses reliable pseudo pairs $\{(b_\varphi(Q_j), Q_j)\}_{j=1}^R$ if the likelihood of generating  $\hat{Q}_j$ exceeds 60\%.
Line 9 optimizes the parameter $\theta$ of $f_{\theta}$ based on pseudo pairs $\{(b_{\varphi}(Q_j), Q_j) \}_{j=1}^R$. 
To ensure that the performance after multiple iterations remains consistent with the original training dataset $\mathcal{D}$, we further fine-tune the forward model $f_{\theta}$ on $\mathcal{D}$ during the final iteration in lines 11-13.
Line 15 employs $f_{\theta}$ to generate the pseudo question $f_\theta(G_i)$ about $G_i$, on which $N$ pseudo pairs $\{(f_\theta(G_i), G_i)\}_{i=1}^N$ are constructed. 
Line 16 sifts out $S$ pseudo pairs $\{(f_\theta(G_i), G_i)\}_{j=1}^S$ according to semantic relevance and diversity of $f_\theta(G_i)$ and $q_i$, which guarantees that they are similar but different.
Line 17 updates the parameter $\varphi$ of $b_{\varphi}$ on $S$ reliable pseudo pairs $\{(f_{\theta}(G_i), G_i) \}_{i=1}^S$.

%% file: experiment.tex
\section{Experimental Evaluation}
\label{sec:experiment}

\subsection{Experimental Settings}
\vpara{Datasets.}
We evaluate our proposed approach on two widely used benchmark datasets WebQuestions (WQ) and PathQuestions (PQ)~\cite{PQ}.
Specifically, WQ combines 22,989 instances from WebQuestionsSP~\cite{WQ} and ComplexWebQuestions~\cite{CWQ}, which are further divided into 18989/2000/2000 for training/validating/testing.
PQ contains 11,793 instances that are partitioned into 9793/1000/1000 for training/validating/testing.

\begin{table*}[ht]
\centering
\newcolumntype{?}{!{\vrule width 1pt}}

\renewcommand\arraystretch{1.1}
\captionsetup{font=footnotesize}
\scalebox{0.7}{
\begin{tabular}{@{ }c@{ }?@{ }cccc@{ }?@{ }cccc@{ }?@{ }cccc@{}}
\toprule
\multirow{2}{*} {\textbf{Model}} &
\multicolumn{4}{@{}c@{}?@{ }}{\textbf{Top-3 Questions}} &
\multicolumn{4}{@{}c@{}?@{ }}{\textbf{Top-5 Questions}} &
\multicolumn{4}{@{}@{}c}{\textbf{Top-10 Questions}} \\
& \textbf{simCSE}& \textbf{BLEU-1}& \textbf{Diverse@3} & \textbf{Dist-1}& \textbf{simCSE}& \textbf{BLEU-1}& \textbf{Diverse@5} & \textbf{Dist-1}& \textbf{simCSE}& \textbf{BLEU-1}& \textbf{Diverse@10} & \textbf{Dist-1}\\
\midrule
T5 & \underline{87.04} & 52.35 & 22.50 & 34.67 & \underline{86.75}&52.30&25.57 &23.67 & \underline{86.16} & 52.47 &29.80 &14.67\\
BART &85.02&\underline{58.19}&18.98&33.79&84.65&\underline{57.33}&22.42&23.04&84.12&\underline{56.61}&26.79&14.27\\
\midrule
T5+P & 85.58 & 48.61 & 24.81 & 36.82 &85.44&49.50&27.66 &25.36 & 85.24 &50.41&31.10 & 15.37\\
B+P &\textbf{89.28}&\textbf{64.74} &21.52&36.64&\textbf{89.04}&\textbf{64.64}&23.77&24.26&\textbf{88.52}&\textbf{63.99}&26.77&14.32\\
\midrule
Davinci003 & 77.06 & 39.33 & \underline{28.14} & 38.95 
&76.86&39.32&30.18 &27.27 &76.94 &39.38 &31.46 & 16.97\\
ChatGPT & 77.17 & 33.58 &\textbf{29.87} & \underline{39.59}&77.18&33.59&\textbf{32.04} &\underline{28.68} &77.25 &33.75 & \underline{34.38} & \underline{18.04}\\
\midrule
Ours &80.97&47.79&26.44&\textbf{41.08}&81.02&47.78&\underline{31.22}&\textbf{29.26}&80.60&47.85&\textbf{35.99}&\textbf{19.16}\\

\bottomrule
\end{tabular}
}
\caption{Overall evaluation on PQ (\%).}
\label{tb:overall_PQ}
\end{table*}

\begin{table*}[ht]
\centering
\newcolumntype{?}{!{\vrule width 1pt}}
\renewcommand\arraystretch{1.1}
\captionsetup{font=footnotesize}
\scalebox{0.7}{
\begin{tabular}{@{}c@{ }?@{ }cccc@{ }?@{ }cccc@{ }?@{ }cccc@{}}
\toprule
\multirow{2}{*} {\textbf{Model}} &
\multicolumn{4}{@{}c@{}?@{ }}{\textbf{Top-3 Questions}} &
\multicolumn{4}{@{}c@{}?@{ }}{\textbf{Top-5 Questions}} &
\multicolumn{4}{@{}@{}c}{\textbf{Top-10 Questions}} \\
& \textbf{simCSE}& \textbf{BLEU-1}& \textbf{Diverse@3} & \textbf{Dist-1}& \textbf{simCSE}& \textbf{BLEU-1}& \textbf{Diverse@5} & \textbf{Dist-1}& \textbf{simCSE}& \textbf{BLEU-1}& \textbf{Diverse@10} & \textbf{Dist-1}\\
\midrule
T5 & 75.80 & 42.11 & 21.36 & 41.19 & 75.88 & 42.51 & 24.14 & 28.21 & 75.83 & 42.85 & 28.02 & 17.13\\
BART & \underline{81.06} & \underline{48.67} & 24.57 & 41.32 & \underline{81.09}&\underline{48.72}&27.31 & 28.37 & \underline{80.92} & \underline{48.46} & 31.35 & 17.61\\
\midrule
T5+P & 77.97 & 42.76 & 22.42 & 42.27 & 78.04 &43.10&25.53 & 29.26 & 78.04 &43.48 & 29.78 & 17.92\\
B+P & \textbf{81.47} & \textbf{49.08} &24.51 & 41.20 & \textbf{81.26}&\textbf{48.82}&26.95 & 28.36 & \textbf{81.08} &\textbf{48.65} & 30.76 & 17.57 \\
\midrule
Davinci003 & 71.68 & 33.62 & 24.32 & \underline{42.95}& 71.75&33.75& 26.51&\underline{30.50} &71.68&33.61 &29.21 & \underline{19.10}\\
ChatGPT & 74.54 &33.21 &\textbf{28.88} & \textbf{42.96}& 74.38 &33.14&\textbf{31.38} &\textbf{30.82} & 74.40&33.09 & \textbf{33.81} & \textbf{19.47} \\
\midrule
Ours& 79.83 & 46.93 & \underline{25.64} & 42.53 & 79.73&46.82&\underline{28.56} &29.75 & 79.46 & 46.60 & \underline{32.55} & 18.72\\
\bottomrule
\end{tabular}
}
\caption{Overall evaluation on WQ (\%).}
\label{tb:overall_WQ}
\end{table*}

\vpara{Evaluation Metrics.}
We assess the generated questions from two aspects: \textbf{relevance} and \textbf{diversity}.
For each instance, we
evaluate the top-3, top-5, and top-10 generated questions respectively. 
We assess the relevance of generated questions in terms of semantics (\emph{i.e.}, the meaning of the text).
Specifically, we adopt simCSE~\cite{SimCSE} to calculate sentence-level semantic relevance between the generated question and the ground truth.
Besides, we also report the token-level relevance using BLEU~\cite{bleu}, which computes the ratio of the common n-grams between the generated question and the ground truth question.
We evaluate the diversity of generated questions using our proposed $Diverse@k$, which measures the diversity among top-k generated questions for each instance while ensuring their relevance to the ground truth.
In addition, we also report Distinct-n~\cite{Distinct}  (abbreviated as Dist-n), which calculates the proportion of unique n-grams in the generated question.

\vpara{Baselines.} 
We compare two types of baselines: pre-trained language models-based (PLMs-based) and large language models-based (LLMs-based). 
Among PLMs-based baselines, 
\textbf{T5}~\cite{T5} and \textbf{BART}~\cite{BART}, the state-of-the-art PLMs for text generation, are fine-tuned for KBQG. Concretely, we linearize each subgraph $G_i$ into a triple-based sequence and then feed the sequence into BART and T5 to generate top-3, top-5, and top-10 questions.
\textbf{T5+Paraphrase} and \textbf{BART+Paraphrase} (abbreviated as \textbf{T5+P} and \textbf{B+P}) are T5 and BART trained on the original and the paraphrased questions.
Specifically, we first apply a popular paraphrase model, \emph{i.e.}, t5-large-paraphraser-diverse-high-quality, to paraphrase the ground truth questions and then create three paraphrased questions for each ground truth question. Then, we fine-tune T5 and BART using these paraphrased and original questions. 
For LLMs-based baselines, we compare two strong baselines, \emph{i.e.}, \textbf{ChatGPT}\footnote{https://openai.com/blog/chatgpt} and text-davinci-003~\cite{InstructGPT} (abbreviated as \textbf{Davinci003}).

\vpara{Pre-training.} We use  BART-base to instantiate the forward model $f_\theta$ and the backward model $b_{\varphi}$.
For pre-training them, we set the learning rate as 5e-5, the batch size as 8, and the maximum epochs as 20. 

\vpara{Fine-tuning.}
We iteratively fine-tune the forward model $f_\theta$ and the backward model $b_{\varphi}$ with the same training settings as the pre-training process, but with a reduced maximum number of training epochs. 
The two models are trained on the generated reliable pseudo pairs  $\{(b_\varphi(Q_j), Q_j)\}_{j=1}^R$ and  $\{(f_{\theta}(G_i), G_i) \}_{i=1}^S$, respectively, using the pre-trained BART-base as the backbone.

\vpara{Code Implementation.} 
We implement our method using Pytorch and conduct all experimental evaluations on a server. This server is configured with a single Nvidia RTX A6000 GPU (48 GB) and equipped with 256 GB memory.
Our code is available at GitHub\footnote{\href{https://github.com/RUCKBReasoning/DiversifyQG}{https://github.com/RUCKBReasoning/DiversifyQG}}.

\subsection{Overall Evaluation}
\label{exp:overall}
Table~\ref{tb:overall_PQ} and Table~\ref{tb:overall_WQ} report the overall evaluation results on PQ and WQ respectively.
Bold formatting denotes the best results while underlining signifies the second best.
According to the results, we conclude that: \textbf{(1) Injecting paraphrased questions enhances both the diversity and relevance of the KBQG task.} T5+P and B+P exhibit greater diversity on the PQ dataset compared to models trained directly on the original dataset, while also achieving higher relevance on the WQ dataset. This suggests that paraphrasing-based fine-tuning introduces richer semantic patterns and expressive variations, enabling the generation of more diverse and contextually relevant questions.
\textbf{(2) Our approach  surpasses PLMs-based baselines in diversity and  matches or even outperforms LLMs-based baselines, which demonstrates the effectiveness of leveraging external natural questions.}
Although the two PLMs-based enhanced baselines considering paraphrased questions obtain better performance than their base version, their performance is constrained by the limited ability of the paraphrase model.
Alternatively, our approach introduces external questions that cover a much broader range of semantic patterns and expressions. 
Furthermore, we note that our method generally outperforms Davinci003 in diversity and is comparable to ChatGPT.  
LLMs (such as ChatGPT) inherently possess richer semantic knowledge compared to PLMs (like BART, the backbone of our approach), leading to higher diversity scores. However, our method surpasses Davinci003, highlighting the effectiveness of our approach.
\textbf{(3) Our approach achieves comparable performance to PLMs-based baselines in relevance and surpasses LLMs-based baselines.}
We observe that our approach achieves slightly lower semantic scores in terms of simCSE than the best PLMs-based baselines, but outperforms LLMs-based baselines. 
Despite the difference between our method and the best baseline, the simCSE scores of our method already meet the relevance criterion.
SimCSE scores of our approach are all greater than 75\%, which indicates the generated questions are very relevant to the gold questions. 
We conduct an experiment to verify this fact.
Specifically, we randomly select 1000 question pairs from Quora Question Pairs\footnote{\scriptsize{http://qim.fs.quoracdn.net/quora\_duplicate\_questions.tsv}}, with each pair annotated as semantically relevant.
We then calculate the average simCSE score among these question pairs, and the result is 70.33\%. 
In fact, the ideal model should excel in diversity metrics while maintaining balanced relevance scores. Our goal is to enhance diversity while ensuring relevance remains within an acceptable range.

\begin{table}[!t]
\centering
\newcolumntype{?}{!{\vrule width 1pt}}
\renewcommand\arraystretch{1.0}
\captionsetup{font=footnotesize}
\scalebox{0.8}{
\begin{tabular}{@{}c@{ }?ccc@{}}
\toprule
\multirow{2}{*} {\textbf{Model}} &
\multicolumn{2}{c}{\textbf{WQ}} 
\\
& \textbf{Diverse@3}&\textbf{simCSE}\\
\midrule
 Ours & \textbf{25.64} & \textbf{79.83}\\
Ours (w/o ss\_f) & 23.26 & 78.28\\
Ours (w/o ss\_b)&25.22 & 79.70\\
\bottomrule
\end{tabular}
}
\caption{Ablation studies of two selection strategies (\%).}
\label{tab:ablation}
\end{table}

\subsection{Ablation Studies}
\label{sec: ablation}
To further investigate the effectiveness of two selection strategies, this section removes one selection strategy during the training of the forward model and another during the training of the backward model to demonstrate the impact of each selection strategy on model performance.

\subsubsection{\textbf{Effect of Selection Strategy on $f_\theta$}}
To demonstrate the effectiveness of our proposed selection strategy in the forward model $f_\theta$, we remove it (\emph{i.e.}, w/o ss\_f) and use all pseudo pairs to train the forward model. 
Table~\ref{tab:ablation} reports the results in diversity and semantics.
We observe that removing the strategy results in relative declines of 9.28\% in $Diverse@3$ and 1.94\% in simCSE.
These results indicate that pseudo pairs dissimilar to the training data can be noisy and hurt the performance of the forward model.

\subsubsection{\textbf{Effect of Selection Strategy on $b_{\varphi}$}}
We explore whether reliable pseudo pairs in training the backward model $b_{\varphi}$ can improve the performance.
We delete the selection strategy for choosing relevant and diverse pseudo pairs (\emph{i.e.}, w/o ss\_b) but utilize all pseudo pairs to fine-tune the backward model. 
According to the results shown in Table~\ref{tab:ablation},
we observe that removing the selection strategy leads to relative declines of 1.64\% in $Diverse@3$ and 0.16\% in simCSE.
These results indicate that reliable pseudo pairs can improve the performance of the backward model.

\subsection{Human Evaluation}
\label{sec:human evaluation}
We randomly choose 50 instances $\mathcal{S}_{50} = \{(G_i,q_i)\}_{i=1}^{50}$ from the test set of the WQ dataset and then evaluate whether top-3 and top-5 generated questions for each instance have different surface forms while ensuring their relevance to the ground truth.

Concretely, we first invite three graduate students to score the relevance between generated questions and the ground truth.
Then the three students judge the diversity of the generated questions for each instance and average the diversity.
Finally, we average the scores of three students for our approach and two PLMs-based baselines BART and B+P. 
The diversity and relevance are scored on a five-point Likert scale, where 1-point indicates poor diversity and relevance, and 5-point represents perfect diversity and relevance. 
Table~\ref{tb:human_evaluation} reports the results, which shows that our approach can produce more diverse questions than other baselines while achieving respectable performance in terms of relevance with the baselines.
Additionally, we use the Pearson correlation coefficient to evaluate the correlation between $Diverse@k$ and human evaluation. Table~\ref{tb:human_evaluation} reports the result of the Pearson correlation.
We observe that our devised metric $Diverse@k$ is highly consistent with human evaluation, which demonstrates its rationality.

\begin{table}[!t]
\centering
\newcolumntype{?}{!{\vrule width 1pt}}
\renewcommand\arraystretch{1.0}
\captionsetup{font=footnotesize}
\scalebox{0.8}{
\begin{tabular}{@{ }c@{ }?cc?cc@{}}
\toprule
\multirow{2}{*} {\textbf{Model}} &
\multicolumn{2}{c?}{\textbf{Top-3 Questions}} &
\multicolumn{2}{c}{\textbf{Top-5 Questions}}

\\
& \textbf{Diversity}& \textbf{Relevance}& \textbf{Diversity}& \textbf{Relevance}\\
\midrule
BART & 3.67 & 4.05 &3.85 &  4.02\\
B+P & 3.45 & \textbf{4.25} & 3.56 &  \textbf{4.18}\\
Ours & \textbf{3.98} & 3.96 & \textbf{4.21} & 3.89 \\
\midrule
Pearson & 0.921 &- & 0.935 & -\\
\bottomrule
\end{tabular}
}
\caption{Human evaluation results on WQ.}
\label{tb:human_evaluation}
\end{table}

%% file: related.tex
\section{Related Work}
\hide{Our work is closely related to diversifying question generation and diversity evaluation metrics. }
\subsection{Diversifying Question Generation.}
Recently, PLMs-based methods~\cite{KGPT, guo2024survey, DSM, JointGT, AutoQGS, Meta-CQG} have been increasingly applied to automatically generate questions.
Despite the success of PLMs-based models on KBQG, they concentrate on improving the quality of a single generated question but lack diversity.
It is well known that diversity can make the generated question look more natural and human-like. 
Currently, diversifying text generation has attracted the interest of researchers and can be broadly categorized into two approaches: model enhancement and data augmentation.
The former mainly concentrates on modifying the model architecture or revising loss functions.
For example, 
\citet{DivHSK} utilize self-attention-based keyword selection to produce headlines that are diverse yet semantically consistent.
\citet{DiversifyingQuestion} use a continuous latent variable to model the content selection process and explicitly model question types using Conditional Variational Auto-encoder (CVAE) to diversify question generation.
\citet{www21} further inject a control algorithm into CVAE to balance the diversity and accuracy of the generated question.
\citet{ClarificationQuestion} 
 design two loss functions to estimate the distribution of keywords in questions, and generate the question based on them.
However, these methods do not apply to our task due to the differences in their settings compared to ours.
In addition, some researchers also explore data augmentation-based methods.
For instance, \citet{dialogueNon} introduce non-conversational text to diversify dialogue generation.
\citet{AskGoodQuestion} create new (source, target) pairs by a simple back translation method to generate human-like questions. 
Our proposed method can be viewed as a data augmentation approach that leverages external natural questions to enhance diversity.
Although \citet{dialogueNon} have also investigated the effectiveness of introducing external data in dialogue generation, our definition of diversity differs from theirs. They consider diversity as the distinction among all the instances as a whole, whereas we focus on the diversity of top-k generated results of each instance while ensuring their relevance to the ground truth. 
Furthermore, we carefully devise two simple and effective reliable pseudo pairs selection strategies on top of the dual model framework.

\subsection{Diversity Evaluation Metrics.}
For diversifying text generation tasks, diversity evaluation is a core step.
Early studies have proposed some popular diversity evaluation metrics, such as Distinct-n~\cite{Distinct} and Self-BLEU~\cite{selfBLEU}.
For Distinct-n, it is a widely-used metric in various generation tasks, such as text generation~\cite{www21, AskGoodQuestion} and story generation ~\cite{storygeneration}.
To be concrete, Distinct-n is calculated as
the number of distinct tokens divided by the total number of tokens, which makes it more like a measure of the duplication of n-grams rather than diversity.
For Self-BLEU, it first computes the BLEU~\cite{bleu} score of all instance pairs and then takes their average.
It is worth noting the smaller the Self-BLEU score, the more diverse it is. 
Obviously, Distinct-n and self-BLEU are inappropriate to measure the diversity explored in this paper since the two metrics focus on the ratio of distinct n-grams.
Meanwhile, the two diversity metrics ignore semantic relevance to the ground truth, which is the key to assessing diversity. 
In view of this, we propose a novel diversity metric called $Diverse@k$, which measures the diversity among multiple generated questions for each instance while ensuring their relevance.

%% file: conclusion.tex
\section{Conclusion}
This work conducts pilot studies on diversifying KBQG.
We rethink the diversity of questions and suppose that diversity should be that questions expressing the same semantics have different forms of expression.
In light of this,
we design a novel diversity evaluation metric $Diverse@k$ that measures the diversity among the top-k generated questions for each instance while guaranteeing relevance to the ground truth.
Furthermore, we propose a dual framework with two simple yet effective selection strategies to generate diverse questions leveraging external natural questions.
Experimental results demonstrate the superiority of our method.

%% file: limitations.tex
\section{Limitations}
In our approach, we introduce external natural questions to diversify question generation, which can generate questions with different expressions, since these natural questions cover a wider range of semantic patterns and expressions.
However, for instances with simple expressions, the paraphrasing-based method may achieve better performance. 
For example, the ground truth is ``\textit{What religion in Australia that influenced Arthur Schopenhauer?}'', the paraphrasing-based approach generates 
``\textit{What \textcolor{blue}{\textbf{faith}} in Australia \textcolor{blue}{\textbf{inspired}} Arthur Schopenhauer?}''. Our method generates  ``\textit{\textcolor{orange}{\textbf{What is the religion}} in Australia that influenced  Arthur Schopenhauer?
}''.
We observe that the paraphrasing-based approach rewrites ``\textbf{religion}'' to ``\textcolor{blue}{\textbf{faith}}'' and rewrites ``\textbf{influenced}'' to ``\textcolor{blue}{\textbf{inspired}}'', but our method only rewrites ``\textbf{What religion}'' to ``\textcolor{orange}{\textbf{What is the religion}}'', because the paraphrasing-based method focuses on words while ours focuses more on the structure of the sentences.
Therefore, when the sentence's expression is not so diverse, the paraphrasing-based method may be well suited. We could study how to improve both word-level and structure-level diversity in the future.

%% file: ack.tex
\section{Acknowledgments}
This work is supported by National Key Research \& Develop Plan (2023YFF0725100) and the National Natural Science Foundation of China (62322214, U23A20299, 62076245, 62072460, 62172424, 62276270). This work is supported by Public Computing Cloud, Renmin University of China. We deeply appreciate the insightful feedback provided by all reviewers.

%% file: appendix.tex
\section{Appendix}

\subsection{Case Study}
\label{sec: case}
We present top-3 questions for five instances generated by BART, B+P, and Ours on WQ in Table ~\ref{tb:overall_case}.
Concretely, each approach returns top-3 generated questions, where the various surface forms for each instance are marked with different colors.
We observe that the top-3 questions generated by ours are marked with three colors, but BART and B+P are mainly marked with two colors and a few with three colors.
Based on the results, we conclude that top-3 questions generated by our model are more diverse than the baselines (\emph{i.e.}, BART and B+P) because our approach introduces various external natural questions that cover a much broader range of semantic patterns and expressions so that it can benefit the diversifying question generation. 

\begin{figure}[!t]
\centering 
\includegraphics[width=0.48\textwidth]{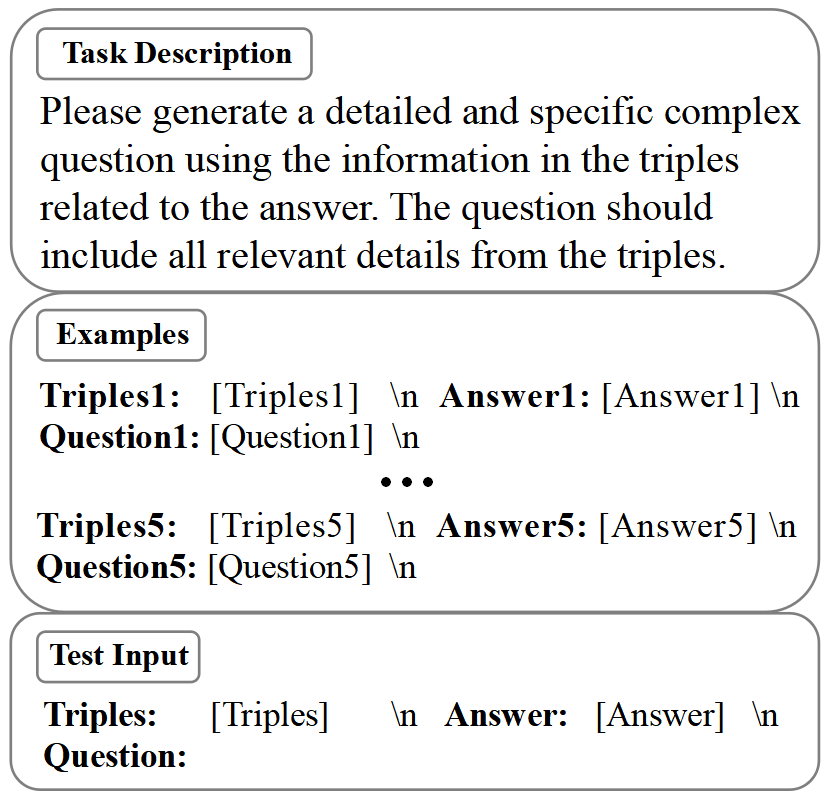}
\caption{The prompt for Large Language Models.}
\label{fig:prompt} 
\end{figure}

\subsection{Prompt for Large Language Models}
\label{sec: prompt}
In our paper, we employ two advanced large language models (LLMs) as baselines, namely ChatGPT and text-davinci-003 (abbreviated as Davinci003).
Our prompt design incorporates three key elements: the task description, illustrative examples, and the test input.
As shown in Figure~\ref{fig:prompt}, the task description meticulously outlines the specifics of the task. 
For each test instance, we choose five representative examples to guide the model's generation.

\begin{table*}[htp]
\centering
\newcolumntype{?}{!{\vrule width 1pt}}
\renewcommand\arraystretch{1.2}
\scalebox{0.7}{
\begin{threeparttable}
\begin{tabular}{@{}c@{ }?@{  }c@{  }?@{  }c@{   }?@{  }c@{  }}
\toprule
\textbf{Ground Truth}& \textbf{BART} & \textbf{B+P}& \textbf{Ours}\\
\midrule
\tabincell{c}{Who is the coach of the team \\owned by Steve Bisciotti?
}  & \tabincell{l}{\textbf{Q1:} \colorbox{LightBlue}{Who is the  coach of the} \\ \colorbox{LightBlue}{team owned by Steve Bisciotti?}\\ \\
\textbf{Q2:} \colorbox{PaleGreen}{Team owner  Steve Bisciotti 's} \\\colorbox{PaleGreen}{sports team is coached by whom?}\\\\
\textbf{Q3:} \colorbox{PaleGreen}{Team owner  Steve Bisciotti 's} \\\colorbox{PaleGreen}{sports team is coached by whom?}
}  & \tabincell{l}{\textbf{Q1:} \colorbox{pink}{Who is the head coach of the} \\\colorbox{pink}{team owned by Steve Bisciotti?}\\ \\
\textbf{Q2:} \colorbox{pink}{Who is the head coach of the} \\\colorbox{pink}{team owned by Steve Bisciotti?}\\\\
\textbf{Q3:} \colorbox{PaleGreen}{Team owner Steve Bisciotti 's} \\\colorbox{PaleGreen}{sports team is coached by whom?}
} & \tabincell{l}{\textbf{Q1:} \colorbox{LightBlue}{Who is the coach of the}\\ \colorbox{LightBlue}{team owned by Steve Bisciotti?}\\\\
\textbf{Q2:} \colorbox{PeachPuff}{Who is the current head} \\ \colorbox{PeachPuff}{coach of the team owned by} \\ \colorbox{PeachPuff}{Steve Bisciotti?}\\\\
\textbf{Q3:} \colorbox{PaleGreen}{Team owner Steve Bisciotti 's}\\ \colorbox{PaleGreen}{sports team is coached by whom?}}\\
\midrule
\tabincell{c}{People from the country with \\the national anthem Lofsöngur \\speak what language?
}  & \tabincell{l}{\textbf{Q1:} \colorbox{Thistle}{Which language is spoken }\\ \colorbox{Thistle}{in the country  that has national} \\ \colorbox{Thistle}{anthem lofsöngur?}\\\\
\textbf{Q2:} What \colorbox{Thistle}{language is spoken} \\ \colorbox{Thistle}{in the country that has national}\\ \colorbox{Thistle}{anthem lofsöngur?}\\\\
\textbf{Q3:} \colorbox{Apricot}{What  language is used} \\ \colorbox{Apricot}{in the country with national}\\ \colorbox{Apricot}{anthem lofsöngur?}
}  & \tabincell{l}{\textbf{Q1:} \colorbox{Beige}{Which language is spoken} \\ \colorbox{Beige}{in the country with the national}\\ \colorbox{Beige}{anthem lofsöngur?}\\\\
\textbf{Q2:} \colorbox{Lavender}{What spoken language is used}\\ \colorbox{Lavender}{in the country with national}\\ \colorbox{Lavender}{anthem lofsöngur?}\\\\
\textbf{Q3:} \colorbox{Lavender}{What spoken language} was\\ \colorbox{Lavender}{used in the country with national}\\ \colorbox{Lavender}{anthem lofsöngur?} 
}& \tabincell{l}{\textbf{Q1:} \colorbox{Lavender}{What spoken language is used}\\ \colorbox{Lavender}{in the country with national}\\ \colorbox{Lavender}{anthem lofsöngur?}\\\\
\textbf{Q2:} \colorbox{Wheat}{People from the country}\\ \colorbox{Wheat}{that has the national anthem} \\ \colorbox{Wheat}{lofsöngur speak what language?}\\\\
\textbf{Q3:} \colorbox{Beige}{Which language is spoken}\\ \colorbox{Beige}{in the country with the national}\\ \colorbox{Beige}{anthem lofsöngur?}
}\\
\midrule
\tabincell{c}{What Canadian religion has a \\religious belief named Mahdi?
}  & \tabincell{l}{\textbf{Q1:} \colorbox{LightBlue}{What religion with religious}\\ \colorbox{LightBlue}{belief named Mahdi is recognized}\\ \colorbox{LightBlue}{in Canada?}\\\\
\textbf{Q2:} \colorbox{LightBlue}{What religion with religious}\\ \colorbox{LightBlue}{belief named Mahdi is recognized}\\ \colorbox{LightBlue}{in Canada?}\\\\
\textbf{Q3:} \colorbox{Khaki}{Which religion with religious}\\ \colorbox{Khaki}{belief in Mahdi is recognized}\\ \colorbox{Khaki}{in Canada?}
}  & \tabincell{l}{\textbf{Q1:} \colorbox{Thistle}{In Canada, what faith Mahdi is}\\ \colorbox{Thistle}{recognised?}\\\\
\textbf{Q2:} \colorbox{Thistle}{In Canada, what faith Mahdi is}\\ \colorbox{Thistle}{recognized?}\\\\
\textbf{Q3:} \colorbox{PeachPuff}{What faith Mahdi is recognized} \\ \colorbox{PeachPuff}{in Canada?}
} & \tabincell{l}{\textbf{Q1:} \colorbox{MistyRose}{What religion with religious}\\ \colorbox{MistyRose}{belief Mahdi is recognized}\\ \colorbox{MistyRose}{in Canada?}\\\\
\textbf{Q2:} \colorbox{LightCyan}{Which of the major religions} \\\colorbox{LightCyan}{of Canada believes in Mahdi?}\\\\
\textbf{Q3:} \colorbox{SandyBrown}{What religion with religious}\\ \colorbox{SandyBrown}{belief Mahdi is in Canada?}
}\\
\midrule
\tabincell{c}{What team with a mascot named\\ K. C. Wolf did Warren \\Moon play for?
}  & \tabincell{l}{\textbf{Q1:} \colorbox{LightYellow}{What team with a mascot} \\\colorbox{LightYellow}{named K. C. Wolf did Warren}\\ \colorbox{LightYellow}{Moon play for?}\\\\
\textbf{Q2:} \colorbox{Lavender}{What team with a mascot} \\ \colorbox{Lavender}{named K. C. Wolf did Warren}\\  \colorbox{Lavender}{Moon play for in 2012?}\\\\
\textbf{Q3:} Which \colorbox{Lavender}{team with a mascot} \\\colorbox{Lavender}{named K. C. Wolf did Warren} \\\colorbox{Lavender}{Moon play for in 2012?}
}  & \tabincell{l}{\textbf{Q1:} \colorbox{LightYellow}{What team with a mascot}\\ \colorbox{LightYellow}{named K. C. Wolf did Warren}\\ \colorbox{LightYellow}{Moon play for?}\\\\
\textbf{Q2:} \colorbox{Bisque}{What team with a K. C.} \\ \colorbox{Bisque}{Wolf mascot did Warren}\\ \colorbox{Bisque}{Moon play for?}\\\\
\textbf{Q3:} \colorbox{LightBlue}{Who did Warren Moon play}\\ \colorbox{LightBlue}{for that has a mascot named} \\ \colorbox{LightBlue}{K. C. Wolf?}
} & \tabincell{l}{\textbf{Q1:} \colorbox{pink}{In what team with a mascot}\\ \colorbox{pink}{named K. C. Wolf did Warren}\\ \colorbox{pink}{Moon play?}\\\\
\textbf{Q2:} \colorbox{Thistle}{Which team with a K. C.}\\ \colorbox{Thistle}{Wolf as a mascot did Warren}\\ \colorbox{Thistle}{Moon play for?}\\\\
\textbf{Q3:} Which \colorbox{Bisque}{team with a K. C.} \\\colorbox{Bisque}{Wolf mascot did Warren Moon}\\ \colorbox{Bisque}{play for?}
}\\

\midrule
\tabincell{c}{What stop motion film featured\\ Miley Cyrus?
}  & \tabincell{l}{\textbf{Q1:} \colorbox{Thistle}{What stop motion movies has}\\ \colorbox{Thistle}{Miley Cyrus been in?}\\\\
\textbf{Q2:} \colorbox{LightBlue}{What stop motion movies} \\\colorbox{LightBlue}{stared Miley Cyrus?}\\\\
\textbf{Q3:} Which \colorbox{LightBlue}{stop motion movies} \\\colorbox{LightBlue}{stared Miley Cyrus?}
}  & \tabincell{l}{\textbf{Q1:} \colorbox{Khaki}{What stop motion movies did}\\ \colorbox{Khaki}{Miley Cyrus appear in?}\\\\
\textbf{Q2:} \colorbox{Gold}{What stop motion movies did}\\ \colorbox{Gold}{Miley Cyrus play in?}\\\\
\textbf{Q3:} \colorbox{DarkSeaGreen}{What stop motion film starred}\\ \colorbox{DarkSeaGreen}{Miley Cyrus?}
} & \tabincell{l}{\textbf{Q1:} \colorbox{LightSteelBlue}{What movies that were}\\ \colorbox{LightSteelBlue}{filmed in stop motion was} \\ \colorbox{LightSteelBlue}{Miley Cyrus in?}\\\\
\textbf{Q2:} \colorbox{Lavender}{What movie featured Miley}\\ \colorbox{Lavender}{Cyrus and was filmed in}\\ \colorbox{Lavender}{stop motion?}\\\\
\textbf{Q3:} \colorbox{Bisque}{What stop motion movies did}\\ \colorbox{Bisque}{Miley Cyrus play?}}\\
\bottomrule
\end{tabular}
\end{threeparttable}
}
\caption{Comparison of top-3 generated questions on WQ, where the various surface forms are marked in different colors.}
\label{tb:overall_case}
\end{table*}

%% file: main.bbl
\begin{thebibliography}{36}
\expandafter\ifx\csname natexlab\endcsname\relax\def\natexlab#1{#1}\fi

\bibitem[{Chen et~al.(2020)Chen, Su, Yan, and Wang}]{KGPT}
Wenhu Chen, Yu~Su, Xifeng Yan, and William~Yang Wang. 2020.
\newblock {KGPT:} knowledge-grounded pre-training for data-to-text generation.
\newblock In \emph{Proceedings of the 2020 Conference on Empirical Methods in Natural Language Processing, {EMNLP} 2020}, pages 8635--8648.

\bibitem[{Chen et~al.(2023)Chen, Wu, and Zaki}]{chen2020toward}
Yu~Chen, Lingfei Wu, and Mohammed~J. Zaki. 2023.
\newblock Toward subgraph-guided knowledge graph question generation with graph neural networks.
\newblock \emph{{IEEE} Transactions on Neural Networks and Learning Systems}, pages 1--12.

\bibitem[{Elangovan et~al.(2023)Elangovan, Maurya, Kumar, and Desarkar}]{DivHSK}
Venkatesh Elangovan, Kaushal Maurya, Deepak Kumar, and Maunendra~Sankar Desarkar. 2023.
\newblock Divhsk: Diverse headline generation using self-attention based keyword selection.
\newblock In \emph{Findings of the Association for Computational Linguistics: {ACL} 2023}, pages 1879--1891.

\bibitem[{Gao et~al.(2021)Gao, Yao, and Chen}]{SimCSE}
Tianyu Gao, Xingcheng Yao, and Danqi Chen. 2021.
\newblock Simcse: Simple contrastive learning of sentence embeddings.
\newblock In \emph{Proceedings of the 2021 Conference on Empirical Methods in Natural Language Processing, {EMNLP} 2021}, pages 6894--6910.

\bibitem[{Guan et~al.(2021)Guan, Zhang, Feng, Liu, Ding, Mao, Fan, and Huang}]{storygeneration}
Jian Guan, Zhexin Zhang, Zhuoer Feng, Zitao Liu, Wenbiao Ding, Xiaoxi Mao, Changjie Fan, and Minlie Huang. 2021.
\newblock Openmeva: {A} benchmark for evaluating open-ended story generation metrics.
\newblock In \emph{Proceedings of the 59th Annual Meeting of the Association for Computational Linguistics and the 11th International Joint Conference on Natural Language Processing, {ACL/IJCNLP} 2021}, pages 6394--6407.

\bibitem[{Guo et~al.(2024)Guo, Liao, Li, and Chua}]{guo2024survey}
Shasha Guo, Lizi Liao, Cuiping Li, and Tat-Seng Chua. 2024.
\newblock A survey on neural question generation: Methods, applications, and prospects.
\newblock \emph{arXiv preprint arXiv:2402.18267}.

\bibitem[{Guo et~al.(2022)Guo, Zhang, Wang, Zhang, Li, and Chen}]{DSM}
Shasha Guo, Jing Zhang, Yanling Wang, Qianyi Zhang, Cuiping Li, and Hong Chen. 2022.
\newblock Dsm: Question generation over knowledge base via modeling diverse subgraphs with meta-learner.
\newblock In \emph{Proceedings of the 2022 Conference on Empirical Methods in Natural Language Processing, {EMNLP} 2022}, pages 4194--4207.

\bibitem[{Hu et~al.(2022)Hu, Hayashi, Cho, and Neubig}]{MTACL22}
Junjie Hu, Hiroaki Hayashi, Kyunghyun Cho, and Graham Neubig. 2022.
\newblock {DEEP:} denoising entity pre-training for neural machine translation.
\newblock In \emph{Proceedings of the 60th Annual Meeting of the Association for Computational Linguistics (Volume 1: Long Papers), {ACL} 2022}, pages 1753--1766.

\bibitem[{Jia et~al.(2020)Jia, Zhou, Sun, and Wu}]{AskGoodQuestion}
Xin Jia, Wenjie Zhou, Xu~Sun, and Yunfang Wu. 2020.
\newblock How to ask good questions? try to leverage paraphrases.
\newblock In \emph{Proceedings of the 58th Annual Meeting of the Association for Computational Linguistics, {ACL} 2020}, pages 6130--6140.

\bibitem[{Ke et~al.(2021)Ke, Ji, Ran, Cui, Wang, Song, Zhu, and Huang}]{JointGT}
Pei Ke, Haozhe Ji, Yu~Ran, Xin Cui, Liwei Wang, Linfeng Song, Xiaoyan Zhu, and Minlie Huang. 2021.
\newblock Jointgt: Graph-text joint representation learning for text generation from knowledge graphs.
\newblock In \emph{Findings of the Association for Computational Linguistics: {ACL/IJCNLP} 2021}, pages 2526--2538.

\bibitem[{Kwiatkowski et~al.(2019)Kwiatkowski, Palomaki, Redfield, Collins, Parikh, Alberti, Epstein, Polosukhin, Devlin, Lee, Toutanova, Jones, Kelcey, Chang, Dai, Uszkoreit, Le, and Petrov}]{NaturalQuestions}
Tom Kwiatkowski, Jennimaria Palomaki, Olivia Redfield, Michael Collins, Ankur~P. Parikh, Chris Alberti, Danielle Epstein, Illia Polosukhin, Jacob Devlin, Kenton Lee, Kristina Toutanova, Llion Jones, Matthew Kelcey, Ming{-}Wei Chang, Andrew~M. Dai, Jakob Uszkoreit, Quoc Le, and Slav Petrov. 2019.
\newblock Natural questions: a benchmark for question answering research.
\newblock \emph{Trans. Assoc. Comput. Linguistics}, 7:452--466.

\bibitem[{Lewis et~al.(2020)Lewis, Liu, Goyal, Ghazvininejad, Mohamed, Levy, Stoyanov, and Zettlemoyer}]{BART}
Mike Lewis, Yinhan Liu, Naman Goyal, Marjan Ghazvininejad, Abdelrahman Mohamed, Omer Levy, Veselin Stoyanov, and Luke Zettlemoyer. 2020.
\newblock {BART:} denoising sequence-to-sequence pre-training for natural language generation, translation, and comprehension.
\newblock In \emph{Proceedings of the 58th Annual Meeting of the Association for Computational Linguistics, {ACL} 2020}, pages 7871--7880.

\bibitem[{Li et~al.(2016)Li, Galley, Brockett, Gao, and Dolan}]{Distinct}
Jiwei Li, Michel Galley, Chris Brockett, Jianfeng Gao, and Bill Dolan. 2016.
\newblock A diversity-promoting objective function for neural conversation models.
\newblock In \emph{Proceedings of the 2016 Conference of the North American Chapter of the Association for Computational Linguistics: Human Language Technologies, {NAACL-HLT} 2016}, pages 110--119.

\bibitem[{Lin and Och(2004)}]{rouge}
Chin-Yew Lin and FJ~Och. 2004.
\newblock Looking for a few good metrics: Rouge and its evaluation.
\newblock In \emph{Ntcir workshop}, pages 1--8.

\bibitem[{Ling et~al.(2020)Ling, Cai, Chen, and de~Rijke}]{DialogueSystem}
Yanxiang Ling, Fei Cai, Honghui Chen, and Maarten de~Rijke. 2020.
\newblock Leveraging context for neural question generation in open-domain dialogue systems.
\newblock In \emph{Proceedings of the Web Conference 2020, {WWW} 2020}, pages 2486--2492.

\bibitem[{Liu et~al.(2021)Liu, Fang, Ding, Li, Wu, and Liu}]{MathWordProblem}
Tianqiao Liu, Qiang Fang, Wenbiao Ding, Hang Li, Zhongqin Wu, and Zitao Liu. 2021.
\newblock Mathematical word problem generation from commonsense knowledge graph and equations.
\newblock In \emph{Proceedings of the 2021 Conference on Empirical Methods in Natural Language Processing, {EMNLP} 2021}, pages 4225--4240.

\bibitem[{Ouyang et~al.(2022)Ouyang, Wu, Jiang, Almeida, Wainwright, Mishkin, Zhang, Agarwal, Slama, Ray, Schulman, Hilton, Kelton, Miller, Simens, Askell, Welinder, Christiano, Leike, and Lowe}]{InstructGPT}
Long Ouyang, Jeffrey Wu, Xu~Jiang, Diogo Almeida, Carroll~L. Wainwright, Pamela Mishkin, Chong Zhang, Sandhini Agarwal, Katarina Slama, Alex Ray, John Schulman, Jacob Hilton, Fraser Kelton, Luke Miller, Maddie Simens, Amanda Askell, Peter Welinder, Paul~F. Christiano, Jan Leike, and Ryan Lowe. 2022.
\newblock Training language models to follow instructions with human feedback.
\newblock In \emph{Proceedings of the 36th Conference on Neural Information Processing Systems, {NeurIPS} 2022}, pages 27730--27744.

\bibitem[{Papineni et~al.(2002{\natexlab{a}})Papineni, Roukos, Ward, and Zhu}]{papineni2002bleu}
Kishore Papineni, Salim Roukos, Todd Ward, and Wei-Jing Zhu. 2002{\natexlab{a}}.
\newblock Bleu: a method for automatic evaluation of machine translation.
\newblock In \emph{Proceedings of the 40th annual meeting of the Association for Computational Linguistics, {ACL} 2002}, pages 311--318.

\bibitem[{Papineni et~al.(2002{\natexlab{b}})Papineni, Roukos, Ward, and Zhu}]{bleu}
Kishore Papineni, Salim Roukos, Todd Ward, and Wei-Jing Zhu. 2002{\natexlab{b}}.
\newblock Bleu: a method for automatic evaluation of machine translation.
\newblock In \emph{Proceedings of the 40th annual meeting of the Association for Computational Linguistics, {ACL} 2002}, pages 311--318.

\bibitem[{Raffel et~al.(2020)Raffel, Shazeer, Roberts, Lee, Narang, Matena, Zhou, Li, and Liu}]{T5}
Colin Raffel, Noam Shazeer, Adam Roberts, Katherine Lee, Sharan Narang, Michael Matena, Yanqi Zhou, Wei Li, and Peter~J. Liu. 2020.
\newblock Exploring the limits of transfer learning with a unified text-to-text transformer.
\newblock \emph{The Journal of Machine Learning Research}, 21(1):5485--5551.

\bibitem[{Shao et~al.(2021)Shao, Wang, Lin, Zhang, Zhang, Ji, and Abdelzaher}]{www21}
Huajie Shao, Jun Wang, Haohong Lin, Xuezhou Zhang, Aston Zhang, Heng Ji, and Tarek~F. Abdelzaher. 2021.
\newblock Controllable and diverse text generation in e-commerce.
\newblock In \emph{Proceedings of the Web Conference 2021, {WWW} 2021}, pages 2392--2401.

\bibitem[{Su et~al.(2020)Su, Shen, Zhao, Zhou, Hu, Zhong, Niu, and Zhou}]{dialogueNon}
Hui Su, Xiaoyu Shen, Sanqiang Zhao, Xiao Zhou, Pengwei Hu, Randy Zhong, Cheng Niu, and Jie Zhou. 2020.
\newblock Diversifying dialogue generation with non-conversational text.
\newblock In \emph{Proceedings of the 58th Annual Meeting of the Association for Computational Linguistics, {ACL} 2020}, pages 7087--7097.

\bibitem[{Talmor and Berant(2018)}]{CWQ}
Alon Talmor and Jonathan Berant. 2018.
\newblock The web as a knowledge-base for answering complex questions.
\newblock In \emph{Proceedings of the 2018 Conference of the North American Chapter of the Association for Computational Linguistics: Human Language Technologies, {NAACL-HLT} 2018}, pages 641--651.

\bibitem[{Wang et~al.(2021{\natexlab{a}})Wang, Zheng, Jiang, and Huang}]{Dialog1}
Yida Wang, Yinhe Zheng, Yong Jiang, and Minlie Huang. 2021{\natexlab{a}}.
\newblock Diversifying dialog generation via adaptive label smoothing.
\newblock In \emph{Proceedings of the 59th Annual Meeting of the Association for Computational Linguistics and the 11th International Joint Conference on Natural Language Processing, {ACL/IJCNLP} 2021}, pages 3507--3520.

\bibitem[{Wang et~al.(2020)Wang, Rao, Zhang, Qin, Tian, and Wang}]{DiversifyingQuestion}
Zhen Wang, Siwei Rao, Jie Zhang, Zhen Qin, Guangjian Tian, and Jun Wang. 2020.
\newblock Diversify question generation with continuous content selectors and question type modeling.
\newblock In \emph{Findings of the Association for Computational Linguistics: {EMNLP} 2020}, pages 2134--2143.

\bibitem[{Wang et~al.(2021{\natexlab{b}})Wang, Lan, and Baraniuk}]{MWPG}
Zichao Wang, Andrew~S. Lan, and Richard~G. Baraniuk. 2021{\natexlab{b}}.
\newblock Math word problem generation with mathematical consistency and problem context constraints.
\newblock In \emph{Proceedings of the 2021 Conference on Empirical Methods in Natural Language Processing, {EMNLP} 2021}, pages 5986--5999.

\bibitem[{Xiong et~al.(2022)Xiong, Bao, Zhao, Wu, and He}]{AutoQGS}
Guanming Xiong, Junwei Bao, Wen Zhao, Youzheng Wu, and Xiaodong He. 2022.
\newblock Autoqgs: Auto-prompt for low-resource knowledge-based question generation from {SPARQL}.
\newblock In \emph{Proceedings of the 31st {ACM} International Conference on Information {\&} Knowledge Management, {CIKM} 2022}, pages 2250--2259.

\bibitem[{Yih et~al.(2016)Yih, Richardson, Meek, Chang, and Suh}]{WQ}
Wen-tau Yih, Matthew Richardson, Christopher Meek, Ming-Wei Chang, and Jina Suh. 2016.
\newblock The value of semantic parse labeling for knowledge base question answering.
\newblock In \emph{Proceedings of the 54th Annual Meeting of the Association for Computational Linguistics, {ACL} 2016}, pages 201--206.

\bibitem[{Zeng and Nakano(2020)}]{zeng2020exploiting}
Jie Zeng and Yukiko~I Nakano. 2020.
\newblock Exploiting a large-scale knowledge graph for question generation in food preference interview systems.
\newblock In \emph{Proceedings of the 25th International Conference on Intelligent User Interfaces Companion}, pages 53--54.

\bibitem[{Zhang et~al.(2022)Zhang, Qiu, Wang, Bai, Li, Jiang, Shen, and Cheng}]{Meta-CQG}
Kun Zhang, Yunqi Qiu, Yuanzhuo Wang, Long Bai, Wei Li, Xuhui Jiang, Huawei Shen, and Xueqi Cheng. 2022.
\newblock Meta-cqg: {A} meta-learning framework for complex question generation over knowledge bases.
\newblock In \emph{Proceedings of the 29th International Conference on Computational Linguistics, {COLING} 2022}, pages 6105--6114.

\bibitem[{Zhang et~al.(2020)Zhang, Kishore, Wu, Weinberger, and Artzi}]{BERTScore}
Tianyi Zhang, Varsha Kishore, Felix Wu, Kilian~Q. Weinberger, and Yoav Artzi. 2020.
\newblock Bertscore: Evaluating text generation with {BERT}.
\newblock In \emph{8th International Conference on Learning Representations, {ICLR} 2020}, pages 1--43.

\bibitem[{Zhang and Zhu(2021)}]{ClarificationQuestion}
Zhiling Zhang and Kenny~Q. Zhu. 2021.
\newblock Diverse and specific clarification question generation with keywords.
\newblock In \emph{Proceedings of the Web Conference 2021, {WWW} 21}, pages 3501--3511.

\bibitem[{Zhou et~al.(2018)Zhou, Huang, and Zhu}]{PQ}
Mantong Zhou, Minlie Huang, and Xiaoyan Zhu. 2018.
\newblock An interpretable reasoning network for multi-relation question answering.
\newblock In \emph{Proceedings of the 27th International Conference on Computational Linguistics, {COLING} 2018}, pages 2010--2022.

\bibitem[{Zhou et~al.(2021)Zhou, Li, and Li}]{Dialog2}
Wangchunshu Zhou, Qifei Li, and Chenle Li. 2021.
\newblock Learning from perturbations: Diverse and informative dialogue generation with inverse adversarial training.
\newblock In \emph{Proceedings of the 59th Annual Meeting of the Association for Computational Linguistics and the 11th International Joint Conference on Natural Language Processing, {ACL/IJCNLP} 2021}, pages 694--703.

\bibitem[{Zhu et~al.(2020)Zhu, Xia, Wu, He, Qin, Zhou, Li, and Liu}]{machine}
Jinhua Zhu, Yingce Xia, Lijun Wu, Di~He, Tao Qin, Wengang Zhou, Houqiang Li, and Tie{-}Yan Liu. 2020.
\newblock Incorporating {BERT} into neural machine translation.
\newblock In \emph{8th International Conference on Learning Representations, {ICLR} 2020}, pages 1--18.

\bibitem[{Zhu et~al.(2018)Zhu, Lu, Zheng, Guo, Zhang, Wang, and Yu}]{selfBLEU}
Yaoming Zhu, Sidi Lu, Lei Zheng, Jiaxian Guo, Weinan Zhang, Jun Wang, and Yong Yu. 2018.
\newblock Texygen: {A} benchmarking platform for text generation models.
\newblock In \emph{41st International {ACM} {SIGIR} Conference on Research {\&} Development in Information Retrieval, {SIGIR} 2018}, pages 1097--1100.

\end{thebibliography}
